\pdfoutput=1

\documentclass[runningheads]{llncs}
\usepackage{graphicx}

\usepackage{tikz}
\usepackage{comment}
\usepackage{amsmath,amssymb} 
\usepackage{color}

\usepackage[accsupp]{axessibility}  

\usepackage{booktabs}
\usepackage{mathtools}
\usepackage{bbm}
\usepackage{dsfont}
\usepackage[inline]{enumitem}
\usepackage[all]{nowidow}
\usepackage{subcaption}
\usepackage{wrapfig}
\usepackage{xspace}
\usepackage{mathdots} 
\usepackage{expl3}
\usepackage[pagebackref=true, colorlinks=true]{hyperref}

\usepackage{lipsum}
\usepackage{tikz}
\usepackage{floatrow}

\usepackage{float}
\floatstyle{plaintop}
\restylefloat{table}

\newcommand*\dif{\mathop{}\!\mathrm{d}}
\DeclarePairedDelimiter{\norm}{\lVert}{\rVert} 

\def\eg{\emph{e.g.} }

\newcommand{\ooi}{OOI\xspace}
\makeatother

\begin{document}
\pagestyle{headings}
\mainmatter
\def\ECCVSubNumber{2145}  

\title{LaTeRF: Label and Text Driven Object Radiance Fields} 

\titlerunning{LaTeRF: Label and Text Driven Object Radiance Fields}
%
\author{Ashkan Mirzaei \and
Yash Kant \and
Jonathan Kelly \and
Igor Gilitschenski}
\authorrunning{A. Mirzaei et al.}
\institute{University of Toronto
}
\maketitle

\begin{abstract}
Obtaining 3D object representations is important for creating photo-realistic simulations and for collecting AR and VR assets. Neural fields have shown their effectiveness in learning a continuous volumetric representation of a scene from 2D images, but acquiring object representations from these models with weak supervision remains an open challenge. In this paper we introduce LaTeRF, a method for extracting an object of interest from a scene given 2D images of the entire scene, known camera poses, a natural language description of the object, and a set of point-labels of object and non-object points in the input images. To faithfully extract the object from the scene, LaTeRF extends the NeRF formulation with an additional `objectness' probability at each 3D point.  Additionally, we leverage the rich latent space of a pre-trained CLIP model combined with our differentiable object renderer, to inpaint the occluded parts of the object. We demonstrate high-fidelity object extraction on both synthetic and real-world datasets and justify our design choices through an extensive ablation study.

\end{abstract}

\section{Introduction}
\label{sec:introduction}
\begin{figure}[t]
  \centering
   \includegraphics[width=1.0\linewidth]{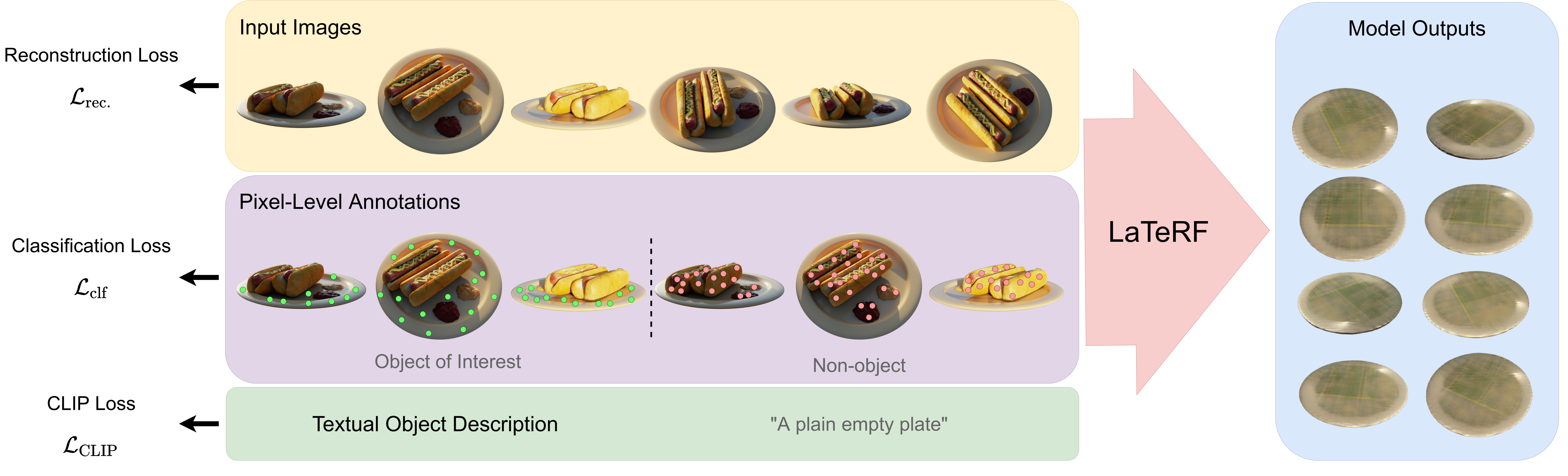}

   \caption{An overview of LaTeRF showing the extraction of a plate. The method takes as input 2D images with camera parameters and poses, a few manually annotated pixel labels corresponding to the object, and labels pointing to the places in the images that do not belong to the object (foreground or background parts). LaTeRF is able to extract the object from the scene while also filling the missing details of the object that are not visible in the input views.}
   \label{fig:method.overview}
\end{figure}

Extracting a partially-occluded object of interest (\ooi) from a 3D scene remains an important problem with numerous applications in computer vision and elsewhere. For example, OOI extraction enables the creation of faithful photo-realistic simulators for robotics, supports the creation of digital content (\eg animated movies), and facilitates the building of AR and VR tools. Although 3D scene reconstruction from 2D images has been extensively explored~\cite{mildenhall2020nerf,jain2021puttingnerfonadiet,wang2021clipnerf,yu2021plenoxels,lin2021barf}, the ability to extract semantically consistent and meaningful objects from complex scenes remains an open challenge.

Early approaches~\cite{Rubinstein_2013_CVPR,6888473} attempted to segment objects from real images, but could not extract their 3D geometry and did not consider the underlying image formation process. Recent work has attempted to disentangle object-centric representations from a scene by learning to decompose the scene into a graph and train object representations at the leaf nodes~\cite{ost2021neural}, or via the use of joint 3D semantic segmentation and neural volumetric rendering~\cite{zhi2021ilabel}. The approach described in~\cite{ost2021neural} is specifically designed for automotive data and is not applicable to object instances from other categories, whereas~\cite{zhi2021ilabel} leads to non-photorealistic object extraction due to training constraints required for real-time deployment. Moreover, none of these works allow for reasoning about occlusions of the \ooi in the scene. Approaches that rely on unsupervised discovery of 3D objects from a single image~\cite{yu2021unsupervised,stelzner2021decomposing} typically require optimizing a non-trivial loss function and do not work well with high-resolution real-world scenes.  
Other approaches~\cite{wu2021dove} require large-scale prior category-level videos that are expensive to collect and might not generalize well to novel categories.


We propose LaTeRF, a novel object extraction method that requires a set of 2D images (with known camera parameters) containing the \ooi mixed with other clutter, a textual description of the \ooi, and a minimal set of pixel annotations distinguishing the object and non-object boundaries. We assign an objectness probability to each point in the 3D space and use this to guide the disentanglement of the \ooi and non-object parts in the scene during training. Once trained, we use an \emph{objectness threshold} to filter points belonging to the \ooi while ignoring non-object points. Moreover, we find that reasoning about occlusion is an important aspect of extracting objects without artifacts (such as holes, or clutter). Thus, we leverage CLIP~\cite{radford2021clip}, a large cross-modal vision-language model, for formulating a loss to fill in the occluded parts using supervision derived from a textual prompt. Combining these two loss functions we can perform high-fidelity object extraction from real-world. An overview of our approach is visualized in Figure~\ref{fig:method.overview}. We evaluate LaTeRF on real and synthetic data and study the effects of individual design choices in extensive ablation studies. We find that, in challenging settings with significant occlusions, LaTeRF leads to $11.19$ and $0.63$ gains in PSNR and SSIM metrics respectively compared to \textit{Mask+NeRF}.

\section{Related Work}
\label{sec:related}
\noindent \textbf{3D object reconstruction:} 
Understanding 3D structure from 2D images has been a long-standing problem in computer vision. Several studies have tried to learn category-specific information from a set of videos~\cite{Henzler_2021_CVPR,wu2021dove} and use this information to generalize to new instances within the same category. Recent work~\cite{kuang2021neroic} captures objects from images with varying illumination and backgrounds, but is unable to inpaint potential missing parts and relies on an object mask for every input image. Neural scene graphs have been utilized for scene decomposition into background and object instances for automotive applications~\cite{ost2021neural}. Another line of work involves recovering the 3D structure of a scene from a single image only and in an unsupervised manner~\cite{yu2021unsupervised,stelzner2021decomposing}, but is limited to simple scenarios and performs poorly in complex scenes. 
Although these methods can recover the 3D shape and appearance of an object with a few 2D images, most rely on training on a large set of either category-specific or general objects; collecting such data can be expensive for certain categories. We aim to overcome the need for prior training and instead employ a minimal set of test-time clues given by a human annotator to extract a photo-realistic object radiance field from the scene. 

\noindent \textbf{Representing scenes with neural fields:} Volumetric rendering of 3D objects and scenes via neural fields has attracted significant attention recently~\cite{tewari2021advancesinneuralrendering}. In the past few years and based on differentiable volume rendering~\cite{tulsiani2017mvsupervision,henzler2019platonicgan}, NeRF-style methods~\cite{mildenhall2020nerf,zhang2020nerf++,niemeyer2021giraffe,rebain2020derf,park2020deformable,yen2020inerf,yu2020pixelnerf} have achieved impressive results in reconstructing high-quality 3D models and novel views learned from 2D inputs of bounded static scenes. The success of NeRF in capturing the details of scenes is due in part to the use of positional encoding~\cite{vaswani2017attentionisallyouneed,gehring2017convolutional} and periodic activations~\cite{sitzmann2019siren,candes1999harmonic,sonoda2017neural} that help to increase the model's capacity. 
The backbone of our method is a neural field\cite{mildenhall2020nerf,lin2020nerfpytorch} that we use to learn the density and view-dependent color of the scene and the object simultaneously based on the camera parameters and input images of the scene containing the object of interest. It is worthwhile to mention that our method is not limited to a specific neural field method and can be extended easily to faster~\cite{liu2020neural,takikawa2021nglod,Jiang:2021} and better-quality NeRFs~\cite{barron2021mipnerf,mueller2022instant}.   

\noindent \textbf{Semantic segmentation in 3D:} Semantic segmentation in 3D has been studied using multi-view fusion-based representations~\cite{armeni_iccv19,mccormac2017semanticfusion,hermans2014dense,ma2017multi,su2015multi,mascaro2021diffuser,vineet2015incremental,zhang2019large} that require only 2D supervision when training, and a separate 3D mesh at testing time, unlike implicit methods like ours. Recently, there have been promising attempts to recover 3D semantic maps from 2D inputs using NeRFs. NeSF~\cite{vora2021nesf} recovers the 3D geometry as density fields from posed RGB images and uses 2D semantic maps to train a semantic field to assign a probability of belonging to each of the semantic classes to every point in the space. Another proposed approach is to extend the NeRF MLP and add a view-independent semantic head to it to get the logits for each of the semantic classes~\cite{zhi2021scenelabelling,zhi2021ilabel}. Our idea is conceptually similar to these, and we propose using binary partitioning of 3D points into two categories (object of interest vs. non-object). In contrast to the previous works, we introduce a differentiable rendering scheme to only render the object of interest using these objectness probabilities, which enables us to fill the visual gaps in the object. 

\noindent \textbf{Language-guided NeRFs:}
In the representation learning~\cite{bengio2013representationlearning} literature, self-supervised approaches~\cite{chen2020simclr,he2020moco,oord2019representation,henaff2020dataefficient} are some of the most compelling settings due to the rather cheap availability of unlabelled data. CLIP~\cite{radford2021clip} is a contrastive-based representation learning method that learns visual and textual feature extraction by looking at a massive set of diverse image-caption pairs scraped from the internet. Following the success of CLIP in finding the similarity of image and text samples, recent works have used it for image generation guided by language~\cite{ramesh2021zeroshottexttoimage}. More recently, fusing CLIP with NeRF has been explored. In~\cite{jain2021puttingnerfonadiet}, CLIP similarity of the rendered scene from different views is utilized to reduce the amount of data needed to train the model. This is based on CLIP's ability to assign similar features to different views of a single object. CLIP-NeRF~\cite{wang2021clipnerf} makes use of the joint image-text latent space of CLIP and provides a framework to manipulate neural radiance fields using multi-modal inputs. Dream field~\cite{jain2021dreamfields} uses the potential of CLIP in finding the similarity between a text and an image and optimizes a NeRF to increase the similarity of its different renders to a text prompt. This way, starting from a phrase, a 3D object closely representing the text is created. 

Motivated by these results, which demonstrate the possibility and benefits of using CLIP alongside neural radiance fields. We leverage the rich multi-modal feature space of the pre-trained CLIP model to give our object extractor semantic information as text. It helps to inpaint the points of the object of interest that are invisible and obscured by other elements in the scene. Notably, we only use CLIP as a good, recent instance of a joint image-language model, and our results can be improved with richer joint embedding functions in the future. The language module of our method is closely related to dream field~\cite{jain2021dreamfields} but we use it to generate 3D objects that are consistent with the ones provided in a scene.


\section{Background}
\label{sec:background}
Neural radiance fields (NeRFs)~\cite{mildenhall2020nerf} use a multi-layer perceptron (MLP) to implicitly capture the geometry and visual appearance of a 3D scene. A scene is encoded as a mapping between the 3D coordinates $x$ and view direction $d$, to a view-dependent color $c$ and view-independent density $\sigma$ using an MLP $f: (x, d) \rightarrow (c, \sigma)$. For simplicity, for every point $x$ and view direction $d$, we write the outputs of the MLP as $c(x, d)$ and $\sigma(x)$, respectively.

Consider a ray $r$ with origin $o$ and direction $d$ characterized by $r(t) = o + td$ with near and far bounds $t_n$ and $t_f$ respectively. Similar to the original NeRF formulation~\cite{mildenhall2020nerf}, the rendering equation to calculate the expected color for the ray $C(r)$ is
\begin{equation}
    \label{eq:nerf.rendering.integral}
    C(r) = \int_{t_n}^{t_f}T(t) \sigma\big(r(t)\big) c\big(r(t), d\big) \dif t,
\end{equation}
where $T(t) = \exp \big(-\int_{t_n}^{t} \sigma(r(s)) \dif s\big)$ is the transmittance. This integral is numerically estimated via quadrature. The interval between $t_n$ and $t_f$ is partitioned into $N$ equal sections, and $t_i$ is uniformly sampled from the $i$-th section. 
In this way, the continuous ray from $t_n$ to $t_f$ is discretized and the estimated rendering integral can be obtained as
\begin{equation}
    \label{eq:discrete.nerf.rendering}
    \hat{C}(r) = \sum_{i = 1}^{N} T_i \big(1 - \exp (-\sigma_i \delta_i)\big) c_i,
\end{equation}
where $T_i = \exp \big( -\sum_{j = 1}^{i - 1} \sigma_j \delta_j \big)$ is the discretized estimate of the transmittance and $\delta_i = t_{i + 1} - t_i$ is the distance between two adjacent sample points along the ray. Note that for the sake of simplicity, $\sigma\big(r(t_i)\big)$ and $c\big(r(t_i), d\big)$ are shown as $\sigma_i$ and $c_i$, respectively. The rendering scheme given by Eq.~\ref{eq:discrete.nerf.rendering} is differentiable, allowing us to train our MLP by minimizing the L2 reconstruction loss between the estimated color $C(r)$ and the ground-truth color $C_\text{GT}(r)$. We use a variant of stochastic gradient descent~\cite{kingma2014method}, and minimize the following loss term:
\begin{equation}
    \label{eq:reconstruction.loss}
    \mathcal{L}_\text{rec.} = \sum_{r \in \mathcal{R}} \norm[\big]{C(r) - C_\text{GT}(r)}^2,
\end{equation}
where $\mathcal{R}$ is a batch of rays sampled from the set of rays where the corresponding pixels are available in the training data. In this paper, we use the reconstruction loss $\mathcal{L}_\text{rec.}$ to ensure the consistency of the extracted object radiance field with respect to that of the original scene. The goal is to make the resulting object look similar to the one represented in the 2D input images of the scene, and not just to generate an object within the same category.

\section{Method}
\label{sec:method}
In this paper, we propose a simple framework to extract 3D objects as radiance fields from a set of 2D input images. 
Built on top of the recent advances in scene representation via neural fields~\cite{tewari2021advancesinneuralrendering,mildenhall2020nerf}, our method aims to softly partition the space into the object and the foreground/background by adding an object probability output to the original NeRF MLP. We leverage pixel-level annotations pointing to the object or the foreground/background and a text prompt expressing the visual features and semantics of the object.

\subsection{Objectness Probability}

We wish to extract an \ooi from a neural radiance field guided by a minimal set of human instructions. Our approach is to reason about the probability of each point in the space being part of the object. Recent work proposed to append a view-independent semantic classifier output~\cite{zhi2021scenelabelling} to the original NeRF architecture. Inspired by this, we extend the NeRF MLP to return an additional objectness score $s(x)$ for every point $x$ in the space. The objectness probability $p(x)$ is obtained from the MLP as
\begin{equation}
    \label{eq:logit.to.probability}
    p(x) = \text{Sigmoid}\big( s(x) \big).
\end{equation}
For an overview of the architecture of the MLP used in LaTeRF please refer to our supplementary material. 

\subsection{Differentiable Object Volume Renderer}
The most straightforward approach to render the object using the NeRF model with the objectness probabilities is to apply a threshold on the probability values and to zero out the density of all points in the space that have an objectness probability less than 0.5. In other words, for every point $x$, the new density function would be $\sigma'(x) = \sigma(x) \mathds{1}\big( p(x) \geq 0.5 \big)$ and the naive object rendering integral would be
\begin{equation}
\label{eq:naive.object.renderer}
C_\text{obj}(r) = \int_{t_n}^{t_f}T_\text{obj}(t) \sigma\big(r(t)\big) \mathds{1}\Big( p\big(r(t)\big) \geq 0.5 \Big) c\big(r(t), d\big) \dif t, 
\end{equation}
where the object transmittance $T_\text{obj}$ is defined as
\begin{equation}
    \label{eq:naive.object.transmittance}
    T_\text{obj}(t) = \exp\bigg( -\int_{t_n}^{t} \sigma\big(r(s)\big)  \mathds{1}\Big( p\big(r(s)\big) \geq 0.5 \Big) \dif s \bigg).
\end{equation}

This approach leads to a hard-partitioning of the scene into two categories: %
\begin{enumerate*}[label=\itshape\arabic*\upshape)]
    \item The extracted object,
    \item background and foreground.
\end{enumerate*}
Although this method works in practice and gives a rough outline of the object, if we apply numerical quadrature to Eq.~\ref{eq:naive.object.renderer} to evaluate the pixel values for the object, the gradients of the resulting pixel colors with respect to the weights of the MLP become zero in most of the domain due to the use of the indicator function. As a result, it is not possible to define a loss function that enforces properties on the rendered images of the object (more on this loss function later).

We fix the aforementioned issue with a minor tweak. The transmittance (density) $\sigma(x)$ can be interpreted as the differential probability of an arbitrary ray passing through $x$ being terminated at the particle at point $x$ in the space. Let $T_x$ represent the event that a ray terminates at point $x$, that is, we have $\sigma(x) = \mathds{P}(T_x)$. In addition, denote the event that location $x$ contains a particle belonging to the \ooi as $O_x$ ($p(x) = \mathds{P}(O_x)$). With this change, the object can be rendered with a new density function $\sigma_\text{obj}(x)$ defined as the joint probability of $T_x$ and $O_x$ by $\sigma_\text{obj}(x) = \mathds{P}(T_x, O_x)$. With the assumption that $T_x$ and $O_x$ are independent, the object density function is:
\begin{equation}
    \label{eq:obj.density.func}
    \sigma_\text{obj}(x) = \mathds{P}(T_x, O_x) = \mathds{P}(T_x) \mathds{P}(O_x) = \sigma(x) p(x).
\end{equation}

Given the above equation, the object rendering integral in Eq.~\ref{eq:naive.object.renderer} can be changed to
\begin{equation}
    \label{eq:object.renderer}
    C_\text{obj}(r) =\\
    \int_{t_n}^{t_f}T_\text{obj}(t) \sigma\big(r(t)\big) p\big(r(t)\big) c\big(r(t), d\big) \dif t,
\end{equation}
and the object transmittance $T_\text{obj}$ in Eq.~\ref{eq:naive.object.transmittance} becomes
\begin{equation}
    \label{eq:object.transmittance}
    T_\text{obj}(t) = \exp\bigg( -\int_{t_n}^{t} \sigma\big(r(s)\big) p\big(r(s)\big) \dif s \bigg).
\end{equation}

Now, similar to the previous quadrature formulation, the integral in Eq.~\ref{eq:object.renderer} is approximated as follows,
\begin{equation}
    \label{eq:discrete.object.rendering}
    \hat{C}_\text{obj}(r) = \sum_{i = 1}^{N} T^\text{obj}_i \big(1 - \exp (-\sigma_i p_i \delta_i)\big) c_i,
\end{equation}
where the discretized object transmittance is $T^\text{obj}_i = \exp \big( -\sum_{j = 1}^{i - 1} \sigma_j p_j \delta_j \big)$ and $p_i = p\big(r(t_i)\big)$. Note that with this new object rendering method, the gradients are smoother and it is possible to use iterative optimization to minimize common loss functions defined over the rendered views of the object. Compared to the previous rendering equation with the indicator function, the new equation can be interpreted as a soft-partitioning of the space where every point belongs to one of the two classes (the object or the foreground/background) and the partitioning is stochastic. 

\subsection{Object Classification using Pixel Annotations}
The first source of `clues' obtained from a human annotator to extract the object is pixel-level annotation information. The user selects a few of the 2D input images of the scene and, for each of them, chooses a few pixels on the \ooi and a few pixels corresponding to the background or the foreground. After collecting these data, for a ray $r$ corresponding to one of the annotated pixels, the label $t(r)$ is defined as $1$ if the pixel is labelled as the object, and $t(r) = 0$ if it is labelled as either foreground or background. The objectness score $S(r)$ is calculated using the classical volume rendering principles:
\begin{equation}
    \label{eq:logit.integral}
    S(r) = \int_{t_n}^{t_f} T(t) \sigma\big(r(t)\big) s\big(r(t)\big) \dif t. 
\end{equation}

The objectness probability for a ray $r$ is obtained as $P(r) = \text{Sigmoid}\big(S(r)\big)$. In our implementation, the numerical estimate of the integral above is determined as
\begin{equation}
    \label{eq:discrete.logit.integral}
    \hat{S}(r) = \sum_{i=1}^{N} T_i \big(1 - \exp(-\sigma_i \delta_i)\big) s_i,
\end{equation}
where $s_i = s\big(r(t_i)\big)$. Consequently, the approximated objectness probability is $\hat{P}(r) = \text{Sigmoid}\big(\hat{S}(r)\big)$ and following~\cite{zhi2021scenelabelling}, the classification loss $\mathcal{L}_\text{clf}$ is defined as
\begin{equation}
    \label{eq:classification.loss}
    \mathcal{L}_\text{clf} = -t(r) \log \hat{P}(r) - \big(1 - t(r)\big) \log \big( 1 - \hat{P}(r) \big).
\end{equation}

This loss function guides the model to tune $s(x)$ to be large for points that belong to the object, and to decrease the value for the rest of the points. The MLP is able to propagate the sparse pixel-level data over the \ooi and the background/foreground~\cite{zhi2021scenelabelling}, resulting in a binary classification of points in the space. However, this loss function alone is not able to reason about the occluded parts of the object, which results in an extracted object with holes and other possible artifacts in areas that are not visible in any of the input images or that  are visible in only a small number of the inputs. 

\subsection{CLIP Loss}
To mitigate the shortcomings of the classification loss in inpainting the unseen parts of the object, we propose the use of a phrase describing the object as the additional input signal. This signal helps to train the MLP to extract the \ooi from the scene with any holes being filled in. The user is asked to describe the object in a few words by pointing out its semantics and appearance. Let $t_\text{text}$ represent this user input.  Subsequently, the CLIP loss is used to make sure that the object, when rendered from random views, is similar to the textual clue. 

The contrastive language-image pre-training (CLIP)~\cite{radford2021clip} model is a multi-modal feature extractor trained on a large dataset of image and caption pairs collected from the Internet. It includes two encoders with a shared normalized latent (output) space.
For an image $I$ and the text $t$ (typically a sentence or phrase), the similarity of their features is proportional to the probability of the text $t$ being associated with $I$. CLIP has a massive and diverse training dataset and has been shown to be useful in zero-shot transfer applications for visual and textual data, for example, object recognition and image synthesis. Recent results suggest that it is applicable to and beneficial for novel view synthesis tasks~\cite{jain2021puttingnerfonadiet,jain2021dreamfields,wang2021clipnerf}. 

The object can be rendered from a random view $v$ pixel-by-pixel using the differentiable object rendering scheme in Eq.~\ref{eq:discrete.object.rendering}, resulting in an image $I_\text{obj}(v)$ based on the current estimate of the object by the model. In order to maximize the similarity between the rendered object $I_\text{obj}(v)$ and the clue phrase $t_\text{text}$, the CLIP loss function $\mathcal{L}_\text{CLIP}$ is defined as
\begin{equation}
    \label{eq:clip.loss}
    \mathcal{L}_\text{CLIP} = - \text{Sim}_\text{CLIP}(I_\text{obj}(v), t_\text{text}),
\end{equation}
where $\text{Sim}_\text{CLIP}(I, t)$ is the similarity of the features of the image $I$ and the text $t$ extracted by the pre-trained CLIP model. This loss is influenced by the recent work~\cite{jain2021dreamfields} that utilizes the CLIP model to generate 3D shapes from text descriptions. Our proposed method begins with a complete scene as a starting point and already has a template of the object, and so uses the CLIP loss to fix the obscured regions and surfaces and to reason about the shape and color of the missing parts. 

\subsection{Training Details}
In summary, the reconstruction loss $\mathcal{L}_\text{rec.}$ is applied to make sure that the final object shape and appearance are consistent with the training images of the scene, while the classification loss $\mathcal{L}_\text{clf}$ is used to guide the model to find the \ooi in the scene and to resolve any potential ambiguities and redundancies that may occur when only using the text query. Meanwhile, the CLIP loss $\mathcal{L}_\text{CLIP}$ facilitates the inpainting of potential holes and occluded parts of the object through a high-level description of semantics and appearance. The final loss function $\mathcal{L}$ used to train the model is
\begin{equation}
    \label{eq:loss.function.overall}
    \mathcal{L} = \mathcal{L}_\text{rec.} + \lambda_\text{clf} \mathcal{L}_\text{clf} + \lambda_\text{CLIP} \mathcal{L}_\text{CLIP},
\end{equation}
where the constants $\lambda_\text{clf}$ and $\lambda_\text{CLIP}$ are the hyperparameters of the model. Having both $\lambda_\text{clf}$ and $\lambda_\text{CLIP}$ set to zero (or close to zero) yields a reproduction of the whole scene without any object extraction.

\section{Experiments}
\label{sec:experiments}
Training an object radiance field using LaTeRF can be done with either $360^\circ$ inward-facing posed 2D images or forward-facing posed views of the scene, in addition to the visual and textual cues for \ooi selection. Besides qualitative results, we study the consequences of applying the different loss terms, the number of visual cues, and the underlying NeRF representation quality on the quality of the resulting object asset on our synthetic dataset that includes challenging scenes with object occlusions. 

\begin{wrapfigure}{r}{0.45\textwidth}
  \centering
  \includegraphics[width=1\linewidth]{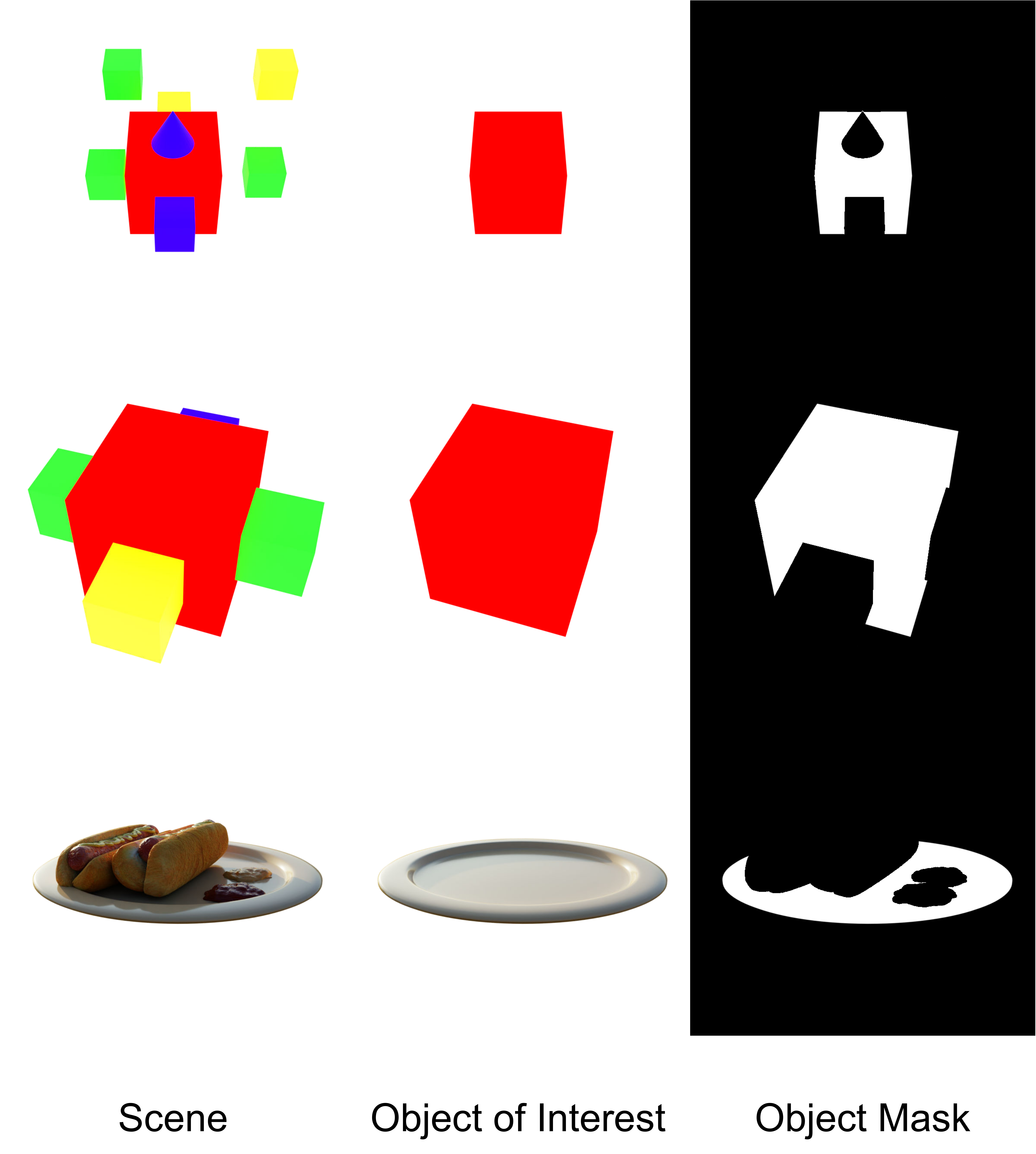}
  \caption{Sample data from 3 of our synthetic scenes.}
  \label{fig:synthetic.data}
\end{wrapfigure}

\noindent\textbf{Datasets} In our experiments, we use a subset of real-world scenes borrowed from NeRD~\cite{boss2021nerd}, scenes that we collected ourselves of real-world objects, and a collection of synthetic scenes full of occlusions that we specifically designed to conduct our quantitative studies. Our synthetic data are rendered using Blender and consists of views of the scene, ground-truth object masks, and ideal extracted objects, as illustrated in Figure~\ref{fig:synthetic.data}. The lighting between different views of most of the scenes provided by NeRD~\cite{boss2021nerd} (scenes which are partly from the British Museum’s photogrammetry dataset~\cite{daniel_pett_gold_cape}) is inconsistent because NeRD was designed for the task of decomposing a scene into the underlying shape, illumination, and reflectance components. As a result, we manually select a subset of the images in these datasets that are roughly similar in terms of lighting and train LaTeRF on them. 

\noindent\textbf{Baseline} To the best of our knowledge, LaTeRF is the only method that is able to extract and inpaint a 3D object from 2D views of a scene, without having a huge category-specific multi-view or video data prior to extraction. As a baseline, we use a NeRF~\cite{mildenhall2020nerf,lin2020nerfpytorch} model (implemented in PyTorch~\cite{NEURIPS2019_bdbca288}) trained on the images of each scene after applying object masks, which we call \textit{Mask+NeRF}. In Mask+NeRF, prior to training, we substitute every pixel marked as non-object in the ground-truth mask with a black pixel and then train a NeRF on the new masked images. 

\noindent\textbf{Metrics} For quantitative comparison of the synthetic dataset, we report the peak signal-to-noise ratio (PSNR) and the structural similarity index measure (SSIM); in both cases, higher is better. 

\subsection{Real-World Scenes}\label{sec:real.world.scenes}

\begin{wrapfigure}{r}{0.47\textwidth}
  \centering
  \includegraphics[width=1\linewidth]{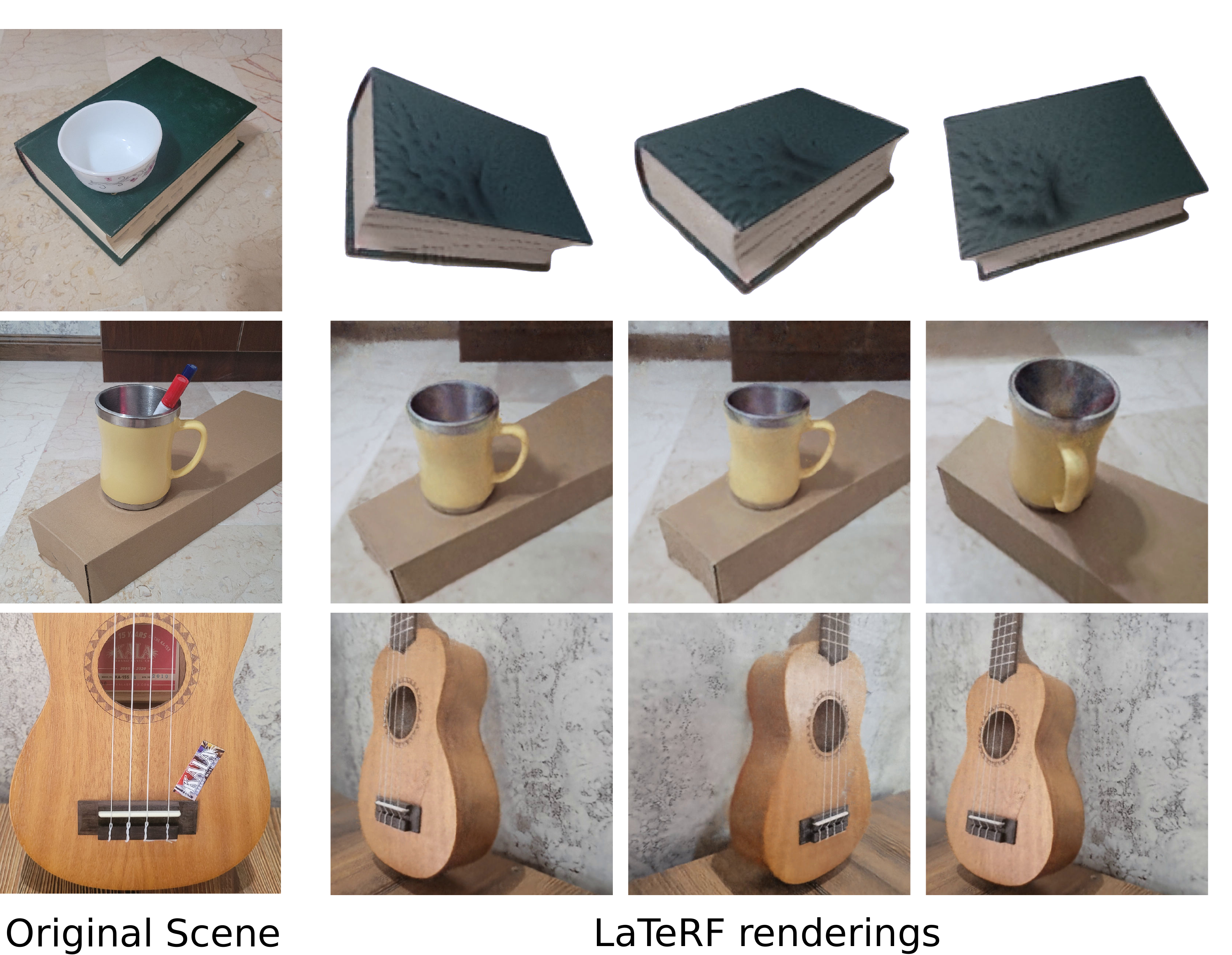}
  \caption{Extracting occluded objects. }
  \label{fig:real.world.objects}
\end{wrapfigure}

\begin{wrapfigure}{r}{0.5\textwidth}
  \centering
  \includegraphics[width=1\linewidth]{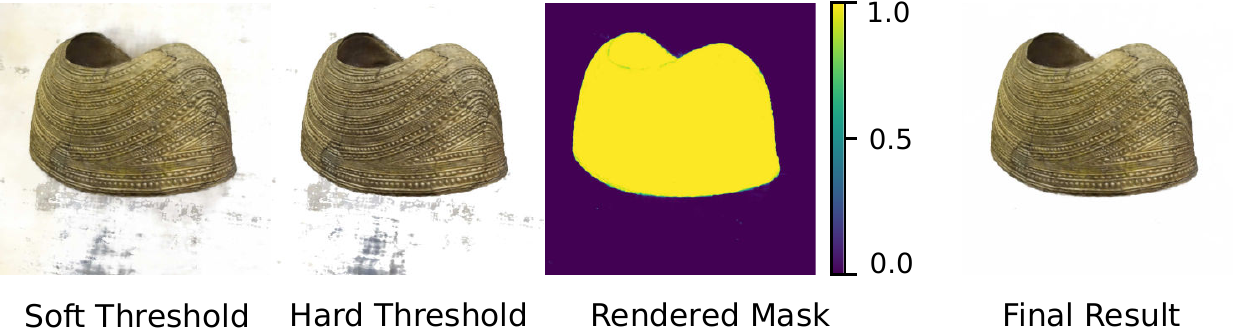}
  \caption{The denoising pipeline.}
  \label{fig:goldcape.pipeline}
\end{wrapfigure}

Qualitative results that demonstrate LaTeRF extracting objects from real-world scenes are shown in Figure~\ref{fig:real.world.objects} (the text queries used in these examples are ”A dark green book”, ”A mug”, and ”A wooden ukulele”, respectively). 
In the real-world scenes with textured and complex backgrounds, we observed that small particles (points with small densities) emerge throughout the scene that are almost invisible from the training views. These particles blend into the background or the \ooi and disappear, but when filtering the object, they become visible. This leads to object renderings full of noise in the background (see the soft threshold results in Figure~\ref{fig:goldcape.pipeline}). In order to remove this noise, we first make the sampled densities along each ray smoother by substituting the density of each point with the average of its density and the density of its direct neighbors; we do this for five steps. After smoothing the densities, most of the small particles, which are abrupt peaks in the value of density, become close to zero. As a result, they can be filtered by applying a hard threshold based on the density and filtering all the points with a value below this threshold. Later, the object mask is rendered by substituting the objectness score instead of the color in Eq.~\ref{eq:discrete.object.rendering} and applying the sigmoid function to the result, while assuming that, along each ray and at an infinitely far point, there is a particle with high density and low objectness score. As a result of this process, the denoised object renderings can be obtained by applying object masks to the noisy images (the result of each step of this denoising pipeline is shown in Figure~\ref{fig:goldcape.pipeline}). Please refer to the supplementary material for more details on the denoising process. 

\subsection{Synthetic Scene Evaluations}

In order to be able to quantitatively evaluate our proposed method in different scenarios, we generate synthetic scenes using Blender, each containing the following information: 
\begin{enumerate*}[label=\arabic*)]
    \item Multi-view $400\times400$ pixel images of the scene that contains different objects including an \ooi, 
    \item the ground-truth images of the \ooi so that we can evaluate the reconstruction quality of our object selection method, 
    \item the mask for the \ooi in each of the $100$ training images of the scene, used instead of manual pixel-level annotations for evaluations, 
    \item and the ground-truth camera intrinsic and extrinsic parameters for each of the input images. 
\end{enumerate*}

Having ground-truth object masks in the dataset makes it easy to automatically sample a certain number of pixels from the object and from the foreground/background to study the effects of pixel labels on the results. 
Sample data from some of the challenging cases of our synthetic dataset can be found in Figure~\ref{fig:synthetic.data}. As shown in the figure, the synthetic scenes have been designed such that parts of the \ooi are obscured with other objects, making the extraction more challenging. 

\subsubsection{Pixel Label Count}

\begin{wraptable}{R}{0.39 \linewidth}
\caption{The effect of the label count on reconstruction quality of the \ooi
}
\begin{center}
\begin{tabular}{lcc}
\hline
Pixel Labels $\#$ & \multicolumn{1}{l}{PSNR$\uparrow$} & \multicolumn{1}{l}{SSIM$\uparrow$} \\ \hline
16,000,000 & \textbf{26.93} & \textbf{0.95} \\
1,600,000 & 26.90 & 0.94 \\
160,000 & 26.48 & 0.90 \\
16,000 & 26.36 & 0.90 \\
1,600 & 26.10 & 0.94 \\
160 & 26.00 & 0.85 \\
16 & 20.81 & 0.86
\end{tabular}
\end{center}
\label{tab:label.count.comparison}
\end{wraptable}

We ablate along the number of pixel labels used during training to demonstrate the effect on reconstruction quality. Since we have the ground-truth object masks in the synthetic dataset, at each step, we simply sample a different number of pixel labels from the mask, based on a uniform distribution, while making sure that the number of positive labels (labels on the \ooi) and negative ones (non-objects) are the same.  Using the sampled subset of the pixel labels, we extract the \ooi from the scene. In order to calculate the reconstruction probability, we compare the results of our method to the ground-truth renderings of the \ooi in the dataset from 30 different viewpoints and report the average PSNR and SSIM. Table~\ref{tab:label.count.comparison} shows the results of this experiment with different numbers of pixel labels. Since each of the $100$ training images has $400\times400$ pixels and the ground-truth object mask includes labels for each of these pixels, there are overall $16,000,000$ labels for each of the scenes. As is evident in Table~\ref{tab:label.count.comparison}, the reconstruction quality of the \ooi stays within the same range when reducing the number of labels from $16,000,000$ to only $160$ labels (an overall of $80$ positive labels and $80$ negative labels that are uniformly distributed across $100$ training images). Note that a total of $160$ labels can easily be obtained within a few minutes by a human annotator and that this is much easier than having someone mask out the entire object for every single training image.

\subsubsection{Importance of Boundary Pixel Labels}

Intuitively, it is clear that not all pixel labels have the same importance. Pixels close to the boundary of the object and non-object portions of each view can help the model to extract the \ooi with sharper and more accurate edges. In this experiment, we show that selecting labels close to the boundaries of the object helps to improve the reconstruction quality. This way, the human annotator can be asked to spend most of their time labelling pixels around the edges. Moreover, we can increase the annotator's brush size (allow the user to label a larger area at a time instead of a single pixel) to allow them quickly annotate the points further from the boundaries and then to focus on the more important boundary labels.

The boundary labels are computed by applying minimum and maximum filters with a kernel size of $3$ on the object masks, reducing the number of labels from $16,000,000$ to less than $500,000$ for our scenes. The pixel labels for this experiment are then uniformly sampled among these boundary pixels. Table~\ref{tab:boundary.label.importance} contains the results with different numbers of boundary labels. As is shown in the table, the reconstruction quality of the model is less affected than the experiment with uniform labels all around the images when reducing the number of boundary labels. The results show an improvement of up to $4.81$ in the PSNR compared to the previous experiment, and this happens when using only $16$ pixel labels. It is worth mentioning that this case has $16$ labels total across all of the input images and not $16$ labels per image. 

\subsubsection{Effects of Different Loss Functions}

As shown in Eq.~\ref{eq:loss.function.overall}, the loss function used to train the main model contains three different parts: 
\begin{enumerate*}[label=\arabic*)]
\item The reconstruction loss $\mathcal{L}_\text{rec.}$, which is the same loss function used in NeRF~\cite{mildenhall2020nerf}, to let the model learn the whole scene, 
\item the classification loss $\mathcal{L}_\text{clf}$, which is defined over the pixel-level annotations to detect the \ooi, 
\item and the CLIP loss $\mathcal{L}_\text{CLIP}$ that guides the model to fill in occluded  portions of the \ooi. 
\end{enumerate*}
In this section, the effect of each of these loss terms on the final reconstruction quality of the model is studied.

\begin{figure}[tb]
\begin{floatrow}
\CenterFloatBoxes
\ttabbox{%
    \begin{tabular}{lcc}
    \hline
    Boundary Labels $\#$ & \multicolumn{1}{l}{PSNR$\uparrow$} & \multicolumn{1}{l}{SSIM$\uparrow$} \\ \hline
    16,000 & \textbf{26.78} & \textbf{0.93} \\
    1,600 & 26.50 & 0.91 \\
    160 & 26.36 & 0.92 \\ 
    16 & 24.82 & 0.90
    \end{tabular}
}{%
  \caption{The effect of the number of boundary labels on the reconstruction quality}\label{tab:boundary.label.importance}
}
\killfloatstyle
\ttabbox{%
    \begin{tabular}{lcc}
    \hline
    Model & \multicolumn{1}{l}{PSNR$\uparrow$} & \multicolumn{1}{l}{SSIM$\uparrow$} \\ \hline
    Mask+NeRF & 15.74 & 0.32 \\
    $\mathcal{L}_\text{rec.}$ & 12.91 & 0.79 \\
    $\mathcal{L}_\text{rec.} + \lambda_\text{clf} \mathcal{L}_\text{clf}$ & 14.10 & 0.82 \\
    $\mathcal{L}_\text{rec.} + \lambda_\text{CLIP} \mathcal{L}_\text{CLIP}$ & 21.95 & 0.84 \\
    LaTeRF (ours) & \textbf{26.93} & \textbf{0.95}
    \end{tabular}
}{%
  \caption{The effect of the loss terms against Mask+NeRF}\label{tab:loss.function.effects}
}
\end{floatrow}
\end{figure}

The reconstruction loss is the main loss enforcing the 3D consistency of the learned scene and \ooi. Thus, it can not be removed and must be present in all of the baselines. Our experiment scenarios involve four cases including: 
\begin{enumerate*}[label=\arabic*)]
    \item $\mathcal{L}_\text{rec.}$, where the classification loss and the CLIP loss are not used. This is similar to the original NeRF~\cite{mildenhall2020nerf} in that there are no clues guiding the model to find the \ooi. 
    \item $\mathcal{L}_\text{rec.} + \lambda_\text{clf} \mathcal{L}_\text{clf}$, where the CLIP loss is ignored and the object is extracted using the pixel labels only. The model has no clue to help reason about missing parts in this case. 
    \item $\mathcal{L}_\text{rec.} + \lambda_\text{CLIP} \mathcal{L}_\text{CLIP}$, where only the text clue is used to `push' the model towards finding the \ooi in the scene. 
    \item $\mathcal{L}_\text{rec.} + \lambda_\text{clf} \mathcal{L}_\text{clf} + \lambda_\text{CLIP} \mathcal{L}_\text{CLIP}$, which is our complete proposed method. 
\end{enumerate*}
We compare the results with the baseline, which is Mask+NeRF. 

\begin{figure*}[t]
  \centering
   \includegraphics[width=1.0\linewidth]{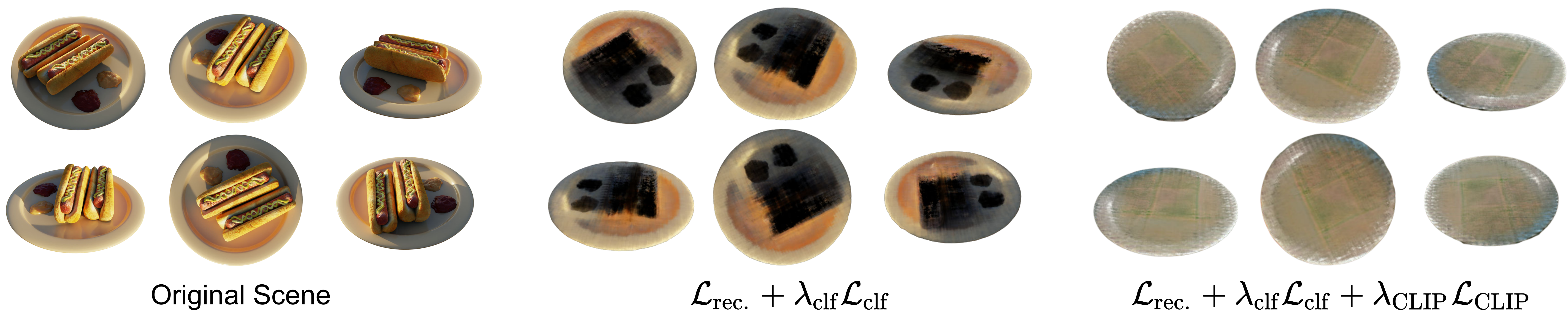}
   \caption{A qualitative representation of the effectiveness of the CLIP loss in filling in object parts that are invisible in the input images. The \ooi is the empty plate. The text prompt used here is \textit{"A plain empty plate"}.  }
   \label{fig:plate.with.different.losses}
\end{figure*}

Table~\ref{tab:loss.function.effects} compares the results of this experiment with the aforementioned baselines. When only the reconstruction loss is used, the semantic head of the MLP can not be trained and the results are unsatisfactory, as expected. The model with the classification loss only is unable to produce results on par with our proposed method, since the synthetic dataset includes objects of interest that are partly covered by other objects (by design), while the CLIP loss is the only semantic clue that is used to fill in the occluded regions. The best result is obtained when using all of the loss terms together; each of them contributes to extracting high-quality objects. 

\begin{wrapfigure}{r}{0.38\textwidth}
  \centering
  \includegraphics[width=1\linewidth]{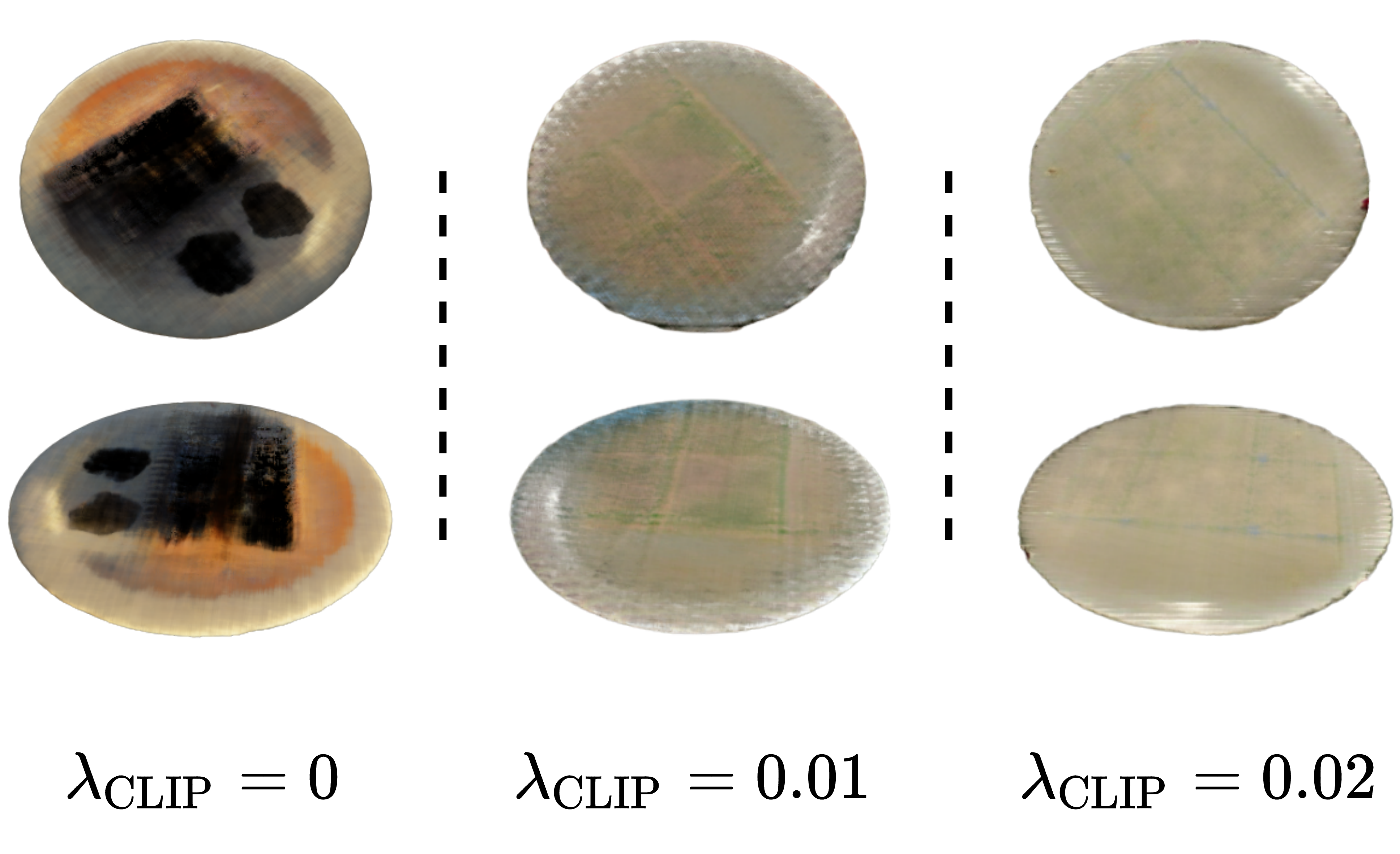}
  \caption{The effect of the weight of the CLIP loss on the extracted plate. }
  \label{fig:clip.loss.plate}
\end{wrapfigure}

Figure~\ref{fig:plate.with.different.losses} visually compares the effect of the presence or absence of the CLIP loss on the rendered object produced. The scene used in this example is a challenging scenario where the goal is to extract the empty plate and remove the contents of the plate. There are various parts of the plate's surface for which no visual information is available in the 2D input images. This example demonstrates that the CLIP loss, making use of a text phrase like \textit{"A plain empty plate"}, is able to reason about the missing area on the surface of the plate and render a plausible empty plate. In addition, the CLIP loss is able to remove the reflectance of the hotdogs on the plate to some extent. However, there are still are limited artifacts on the plate even after applying the CLIP loss. This result is due to the challenging nature of the task, since a large portion of the plate's surface is covered in the input images. As another example, the model has difficulty removing the shadow of the hotdogs from the original plate and tries to complete the shadow instead of removing it. This issue can be mitigated by increasing the CLIP loss, but the result is a plate that is less consistent with the one in the original scene, as shown in Figure~\ref{fig:clip.loss.plate}. Notice that not only the depth of the plate is reduced compared to the one in the original scene when the CLIP weight $\lambda_\text{CLIP}$ is increased to $0.02$; the lighting highlight on the plate surface is also rotated and differs from the original radiance of the plate.  For the case without the CLIP loss (the middle row in Figure~\ref{fig:plate.with.different.losses}), the background has been set to black to better visualize the holes in the plate. 

\subsubsection{Importance of the Number of Input Views}

\begin{wraptable}{R}{0.34 \linewidth}
\caption{The effect of the number of training views for the calculation of $\mathcal{L}_\text{rec.}$}
\begin{center}
\begin{tabular}{l c c}
\hline
$\#$ Views & \multicolumn{1}{l}{PSNR$\uparrow$} & \multicolumn{1}{l}{SSIM$\uparrow$} \\ \hline
100 & \textbf{26.93} & \textbf{0.96} \\
80 & 26.62 & 0.95 \\
60 & 26.50 & 0.92 \\
40 & 26.20 & 0.95 \\
20 & 25.35 & 0.94
\end{tabular}
\end{center}
\label{tab:input.view.number.effect}
\end{wraptable}

The neural volumetric rendering method that is used in LaTeRF is not limited to a certain implementation of NeRF. We argue that with better quality and faster neural rendering approaches, the reconstruction quality and the speed of LaTeRF can be increased. In this experiment, the effect of the quality of the base neural renderer on final object reconstruction quality is studied. In order to mimic the behavior of a lower-quality neural rendering model, we limit the number of training input views fed when calculating the reconstruction loss $\mathcal{L}_\text{rec.}$. The results in Table~\ref{tab:input.view.number.effect} indicate that the detail and quality of the rendered object produced by LaTeRF is closely related to the reconstruction quality of the base NeRF model used to ensure the consistency of the extracted object with the one present in the scene. Consequently, as better volumetric rendering approaches for 3D reconstruction are introduced, LaTeRF will be able to take advantage of them to enable the creation of better object extractors.

\section{Conclusion}
\label{sec:conclusion}
We have presented LaTeRF, a method to extract digital objects via neural fields from 2D input images of a scene, and a minimal set of visual and natural language clues that guide the model to the object of interest. The geometry and color of different points are captured via NeRF, while the objectness probability of scene points is mainly determined via pixel labels provided by the user that identify whether some of the input pixels belong to the object of interest or not. Additionally, a CLIP loss is defined to ensure high similarity between the rendered images of the object and a text prompt that expresses the appearance and semantics of the object, enabling the model to reason about any missing parts of the object. The effectiveness of each of the components of the training scheme is shown through our experiments. However, we observed a need for per-scene fine-tuning of the hyper-parameters $\lambda_\text{clf}$ and $\lambda_{CLIP}$. An additional challenge for LaTeRF is to extract transparent objects or objects with shiny surfaces that reflect other objects. Moreover, LaTeRF naturally comes with the general limitations associated with NeRFs, including the need for per-scene training and data intensity. Additionally, the rendered images used to calculate the CLIP loss had to be downsized to avoid memory shortages. Applying the CLIP loss to higher-resolution images will result in better inpainting.

%
%
\bibliographystyle{splncs04}
\bibliography{egbib}

\begin{thebibliography}{10}
\providecommand{\url}[1]{\texttt{#1}}
\providecommand{\urlprefix}{URL }
\providecommand{\doi}[1]{https://doi.org/#1}

\bibitem{armeni_iccv19}
Armeni, I., He, Z.Y., Gwak, J., Zamir, A.R., Fischer, M., Malik, J., Savarese,
  S.: 3d scene graph: A structure for unified semantics, 3d space, and camera.
  In: ICCV (2019)

\bibitem{barron2021mipnerf}
Barron, J.T., Mildenhall, B., Tancik, M., Hedman, P., Martin-Brualla, R.,
  Srinivasan, P.P.: Mip-nerf: A multiscale representation for anti-aliasing
  neural radiance fields. arXiv  (2021)

\bibitem{bengio2013representationlearning}
Bengio, Y., Courville, A., Vincent, P.: Representation learning: A review and
  new perspectives. In: IEEE transactions on pattern analysis and machine
  intelligence (2013)

\bibitem{boss2021nerd}
Boss, M., Braun, R., Jampani, V., Barron, J.T., Liu, C., Lensch, H.: Nerd:
  Neural reflectance decomposition from image collections. In: ICCV (2021)

\bibitem{candes1999harmonic}
Cand{\`e}s, E.J.: Harmonic analysis of neural networks. Applied and
  Computational Harmonic Analysis  (1999)

\bibitem{chen2020simclr}
Chen, T., Kornblith, S., Norouzi, M., Hinton, G.: A simple framework for
  contrastive learning of visual representations. In: ICML (2020)

\bibitem{gehring2017convolutional}
Gehring, J., Auli, M., Grangier, D., Yarats, D., Dauphin, Y.N.: Convolutional
  sequence to sequence learning. In: ICML (2017)

\bibitem{he2020moco}
He, K., Fan, H., Wu, Y., Xie, S., Girshick, R.: Momentum contrast for
  unsupervised visual representation learning. In: CVPR (2020)

\bibitem{henzler2019platonicgan}
Henzler, P., Mitra, N.J., Ritschel, T.: Escaping plato's cave: 3d shape from
  adversarial rendering. In: ICCV (2019)

\bibitem{Henzler_2021_CVPR}
Henzler, P., Reizenstein, J., Labatut, P., Shapovalov, R., Ritschel, T.,
  Vedaldi, A., Novotny, D.: Unsupervised learning of 3d object categories from
  videos in the wild. In: CVPR (2021)

\bibitem{hermans2014dense}
Hermans, A., Floros, G., Leibe, B.: Dense 3d semantic mapping of indoor scenes
  from rgb-d images. In: ICRA (2014)

\bibitem{henaff2020dataefficient}
Hénaff, O.J., Srinivas, A., Fauw, J.D., Razavi, A., Doersch, C., Eslami,
  S.M.A., van~den Oord, A.: Data-efficient image recognition with contrastive
  predictive coding. In: ICML (2020)

\bibitem{jain2021dreamfields}
Jain, A., Mildenhall, B., Barron, J.T., Abbeel, P., Poole, B.: Zero-shot
  text-guided object generation with dream fields. arXiv  (2021)

\bibitem{jain2021puttingnerfonadiet}
Jain, A., Tancik, M., Abbeel, P.: Putting nerf on a diet: Semantically
  consistent few-shot view synthesis. In: ICCV (2021)

\bibitem{Jiang:2021}
Jiang, G., Kainz, B.: Deep radiance caching: Convolutional autoencoders deeper
  in ray tracing. Computers \& Graphics  (2021)

\bibitem{kingma2014method}
Kingma, D.P., Ba, J.: Adam: A method for stochastic optimization. In: ICLR
  (2015)

\bibitem{kuang2021neroic}
Kuang, Z., Olszewski, K., Chai, M., Huang, Z., Achlioptas, P., Tulyakov, S.:
  {NeROIC}: Neural object capture and rendering from online image collections.
  arXiv  (2022)

\bibitem{lin2021barf}
Lin, C.H., Ma, W.C., Torralba, A., Lucey, S.: Barf: Bundle-adjusting neural
  radiance fields. In: ICCV (2021)

\bibitem{liu2020neural}
Liu, L., Gu, J., Lin, K.Z., Chua, T.S., Theobalt, C.: Neural sparse voxel
  fields. In: NeurIPS (2020)

\bibitem{ma2017multi}
Ma, L., St{\"u}ckler, J., Kerl, C., Cremers, D.: Multi-view deep learning for
  consistent semantic mapping with rgb-d cameras. In: IROS (2017)

\bibitem{mascaro2021diffuser}
Mascaro, R., Teixeira, L., Chli, M.: Diffuser: Multi-view 2d-to-3d label
  diffusion for semantic scene segmentation. In: ICRA (2021)

\bibitem{mccormac2017semanticfusion}
McCormac, J., Handa, A., Davison, A., Leutenegger, S.: Semanticfusion: Dense 3d
  semantic mapping with convolutional neural networks. In: ICRA (2017)

\bibitem{mildenhall2020nerf}
Mildenhall, B., Srinivasan, P.P., Tancik, M., Barron, J.T., Ramamoorthi, R.,
  Ng, R.: Nerf: Representing scenes as neural radiance fields for view
  synthesis. In: ECCV (2020)

\bibitem{mueller2022instant}
M\"uller, T., Evans, A., Schied, C., Keller, A.: Instant neural graphics
  primitives with a multiresolution hash encoding. arXiv  (2022)

\bibitem{niemeyer2021giraffe}
Niemeyer, M., Geiger, A.: Giraffe: Representing scenes as compositional
  generative neural feature fields. In: CVPR (2021)

\bibitem{oord2019representation}
van~den Oord, A., Li, Y., Vinyals, O.: Representation learning with contrastive
  predictive coding. arXiv  (2019)

\bibitem{ost2021neural}
Ost, J., Mannan, F., Thuerey, N., Knodt, J., Heide, F.: Neural scene graphs for
  dynamic scenes. In: CVPR (2021)

\bibitem{park2020deformable}
Park, K., Sinha, U., Barron, J.T., Bouaziz, S., Goldman, D.B., Seitz, S.M.,
  Martin-Brualla, R.: Nerfies: Deformable neural radiance fields. In: ICCV
  (2021)

\bibitem{NEURIPS2019_bdbca288}
Paszke, A., Gross, S., Massa, F., Lerer, A., Bradbury, J., Chanan, G., Killeen,
  T., Lin, Z., Gimelshein, N., Antiga, L., Desmaison, A., Kopf, A., Yang, E.,
  DeVito, Z., Raison, M., Tejani, A., Chilamkurthy, S., Steiner, B., Fang, L.,
  Bai, J., Chintala, S.: Pytorch: An imperative style, high-performance deep
  learning library. In: Wallach, H., Larochelle, H., Beygelzimer, A.,
  d\textquotesingle Alch\'{e}-Buc, F., Fox, E., Garnett, R. (eds.) NeurIPS
  (2019)

\bibitem{daniel_pett_gold_cape}
Pett, D.: {BritishMuseumDH/moldGoldCape: First release of the Cape in 3D}
  (2017). \doi{10.5281/zenodo.344914}

\bibitem{radford2021clip}
Radford, A., Kim, J.W., Hallacy, C., Ramesh, A., Goh, G., Agarwal, S., Sastry,
  G., Askell, A., Mishkin, P., Clark, J., Krueger, G., Sutskever, I.: Learning
  transferable visual models from natural language supervision. In: Meila, M.,
  Zhang, T. (eds.) ICML (2021)

\bibitem{ramesh2021zeroshottexttoimage}
Ramesh, A., Pavlov, M., Goh, G., Gray, S., Voss, C., Radford, A., Chen, M.,
  Sutskever, I.: Zero-shot text-to-image generation. arXiv  (2021)

\bibitem{rebain2020derf}
Rebain, D., Jiang, W., Yazdani, S., Li, K., Yi, K.M., Tagliasacchi, A.: Derf:
  Decomposed radiance fields. In: CVPR (2020)

\bibitem{Rubinstein_2013_CVPR}
Rubinstein, M., Joulin, A., Kopf, J., Liu, C.: Unsupervised joint object
  discovery and segmentation in internet images. In: CVPR (2013)

\bibitem{yu2021plenoxels}
{Sara Fridovich-Keil and Alex Yu}, Tancik, M., Chen, Q., Recht, B., Kanazawa,
  A.: Plenoxels: Radiance fields without neural networks. In: CVPR (2022)

\bibitem{sitzmann2019siren}
Sitzmann, V., Martel, J.N., Bergman, A.W., Lindell, D.B., Wetzstein, G.:
  Implicit neural representations with periodic activation functions. In:
  NeurIPS (2020)

\bibitem{sonoda2017neural}
Sonoda, S., Murata, N.: Neural network with unbounded activation functions is
  universal approximator. Applied and Computational Harmonic Analysis  (2017)

\bibitem{stelzner2021decomposing}
Stelzner, K., Kersting, K., Kosiorek, A.R.: Decomposing 3d scenes into objects
  via unsupervised volume segmentation. arXiv  (2021)

\bibitem{su2015multi}
Su, H., Maji, S., Kalogerakis, E., Learned-Miller, E.: Multi-view convolutional
  neural networks for 3d shape recognition. In: ICCV (2015)

\bibitem{takikawa2021nglod}
Takikawa, T., Litalien, J., Yin, K., Kreis, K., Loop, C., Nowrouzezahrai, D.,
  Jacobson, A., McGuire, M., Fidler, S.: Neural geometric level of detail:
  Real-time rendering with implicit {3D} shapes. In: CVPR (2021)

\bibitem{tewari2021advancesinneuralrendering}
Tewari, A., Thies, J., Mildenhall, B., Srinivasan, P., Tretschk, E., Wang, Y.,
  Lassner, C., Sitzmann, V., Martin-Brualla, R., Lombardi, S., Simon, T.,
  Theobalt, C., Niessner, M., Barron, J.T., Wetzstein, G., Zollhoefer, M.,
  Golyanik, V.: Advances in neural rendering. In: SIGGRAPH (2021)

\bibitem{tulsiani2017mvsupervision}
Tulsiani, S., Zhou, T., Efros, A.A., Malik, J.: Multi-view supervision for
  single-view reconstruction via differentiable ray consistency. In: CVPR
  (2017)

\bibitem{vaswani2017attentionisallyouneed}
Vaswani, A., Shazeer, N., Parmar, N., Uszkoreit, J., Jones, L., Gomez, A.N.,
  Kaiser, L.u., Polosukhin, I.: Attention is all you need. In: NeurIPS (2017)

\bibitem{vineet2015incremental}
Vineet, V., Miksik, O., Lidegaard, M., Nie{\ss}ner, M., Golodetz, S.,
  Prisacariu, V.A., K{\"a}hler, O., Murray, D.W., Izadi, S., P{\'e}rez, P.,
  et~al.: Incremental dense semantic stereo fusion for large-scale semantic
  scene reconstruction. In: ICRA (2015)

\bibitem{vora2021nesf}
Vora, S., Radwan, N., Greff, K., Meyer, H., Genova, K., Sajjadi, M.S.M., Pot,
  E., Tagliasacchi, A., Duckworth, D.: Nesf: Neural semantic fields for
  generalizable semantic segmentation of 3d scenes. arXiv  (2021)

\bibitem{wang2021clipnerf}
Wang, C., Chai, M., He, M., Chen, D., Liao, J.: Clip-nerf: Text-and-image
  driven manipulation of neural radiance fields. arXiv  (2021)

\bibitem{wu2021dove}
Wu, S., Jakab, T., Rupprecht, C., Vedaldi, A.: Dove: Learning deformable 3d
  objects by watching videos. arXiv  (2021)

\bibitem{lin2020nerfpytorch}
Yen-Chen, L.: Nerf-pytorch. \url{https://github.com/yenchenlin/nerf-pytorch/}
  (2020)

\bibitem{yen2020inerf}
Yen-Chen, L., Florence, P., Barron, J.T., Rodriguez, A., Isola, P., Lin, T.Y.:
  {iNeRF}: Inverting neural radiance fields for pose estimation. In: IROS
  (2021)

\bibitem{yu2020pixelnerf}
Yu, A., Ye, V., Tancik, M., Kanazawa, A.: pixelnerf: Neural radiance fields
  from one or few images. In: CVPR (2021)

\bibitem{yu2021unsupervised}
Yu, H.X., Guibas, L.J., Wu, J.: Unsupervised discovery of object radiance
  fields. In: ICLR (2022)

\bibitem{zhang2019large}
Zhang, C., Liu, Z., Liu, G., Huang, D.: Large-scale 3d semantic mapping using
  monocular vision. In: ICIVC (2019)

\bibitem{zhang2020nerf++}
Zhang, K., Riegler, G., Snavely, N., Koltun, V.: Nerf++: Analyzing and
  improving neural radiance fields. arXiv  (2020)

\bibitem{zhi2021scenelabelling}
Zhi, S., Laidlow, T., Leutenegger, S., Davison, A.: In-place scene labelling
  and understanding with implicit scene representation. In: ICCV (2021)

\bibitem{zhi2021ilabel}
Zhi, S., Sucar, E., Mouton, A., Haughton, I., Laidlow, T., Davison, A.J.:
  Ilabel: Interactive neural scene labelling (2021)

\bibitem{6888473}
Zhu, J.Y., Wu, J., Xu, Y., Chang, E., Tu, Z.: Unsupervised object class
  discovery via saliency-guided multiple class learning. IEEE Transactions on
  Pattern Analysis and Machine Intelligence  (2015)

\end{thebibliography}

\newpage
\appendix
\section{Additional Implementation Details}

\begin{figure}[t]
  \centering
   \includegraphics[width=1\linewidth]{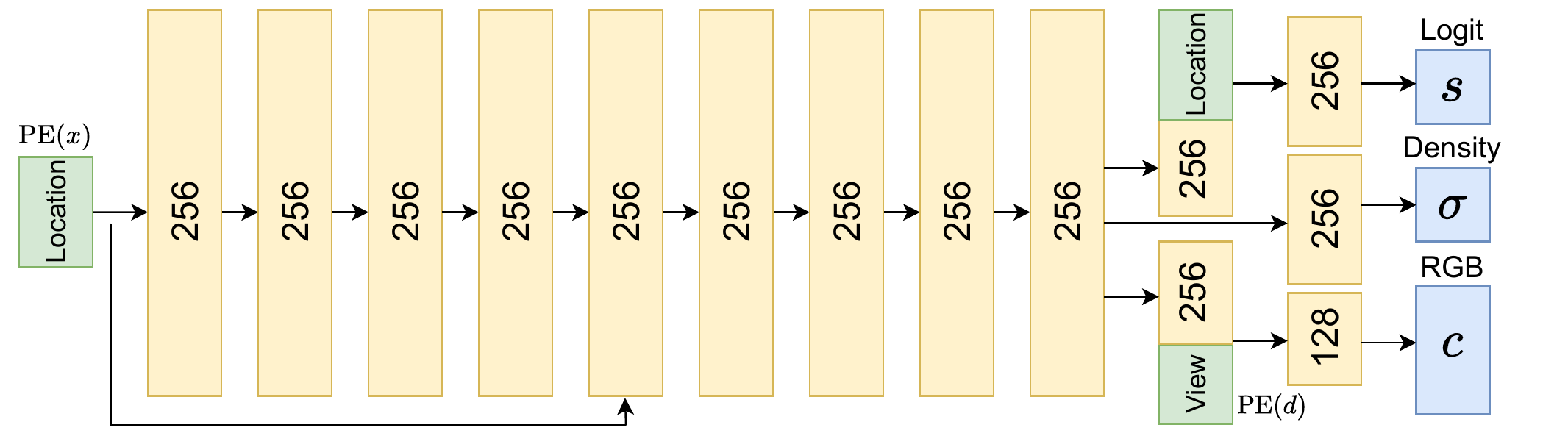}
   \caption{The architecture used for LaTeRF is a multi layer perceptron which inspired by~\cite{zhi2021scenelabelling}, extends the original NeRF model~\cite{mildenhall2020nerf} to contain an additional output $s$ to reason about the probability of different points in the space being part of the \ooi. As evident in the figure, $s$ is independent of the view $d$ and only depends on the location $x$. Following the literature, both location $x$ and view direction $d$ are passed through a positional encoding function ($\text{PE}$). }
   \label{fig:mlp.architecture}
\end{figure}

Our proposed method is implemented in PyTorch~\cite{NEURIPS2019_bdbca288}, and the network is optimized using the Adam optimizer~\cite{kingma2014method} to learn the view-dependent radiance (color) and view-independent densities and objectness probabilities for points in the space. The network is randomly initialized and trained from scratch for each scene individually. Hyperparameters related to the NeRF model and the optimizer are set based on the PyTorch implementation~\cite{lin2020nerfpytorch} in the original NeRF paper~\cite{mildenhall2020nerf}. Training is done on one NVIDIA GeForce RTX 3090 GPU. Due to the high GPU memory usage for the CLIP loss calculations, the object renders used for computing $\mathcal{L}_\text{CLIP}$ are $\frac{1}{3}$ of the size of the original input images, e.g., for $400\times400$ pixel synthetic views, the object is rendered at $133\times133$ pixels and the CLIP loss is defined over the downsized rendering. Moreover, in order to speed up training, the CLIP loss is only calculated every $10$ steps. For test-time object renderings, instead of using the soft-partitioning approach, a hard-thresholding method is used to denoise the background. In the next section, we introduce the details of this denoising mechanism. An overview of the architecture of the MLP with the additional objectness score $s$ as output is shown in Figure~\ref{fig:mlp.architecture}.

For real-world scenes, a minimal user interface is designed to capture the pixel annotations from an end-user. In this procedure, instead of limiting the user to give visual cues pixel-by-pixel, we allow brush size changes to enable the selection of areas corresponding to either the \ooi or non-objects. This allows for quick annotation of points far from the object boundaries, resulting in a better non-object removal and object discovery in the scene. Using dynamic brush sizes, we were able to collect millions of pixel-level annotations in just a few minutes. By being able to change the brush size, the end-user can coarsely label the points that are not close to the boundaries, and then reduce the annotation area as they get closer to the boundaries to finely label the more important pixel-level data (i.e., the boundary of the object and the foreground/background, as shown in the experiments). 

\section{Denoising the Views}

\begin{figure}[t]
  \centering
   \includegraphics[width=1\linewidth]{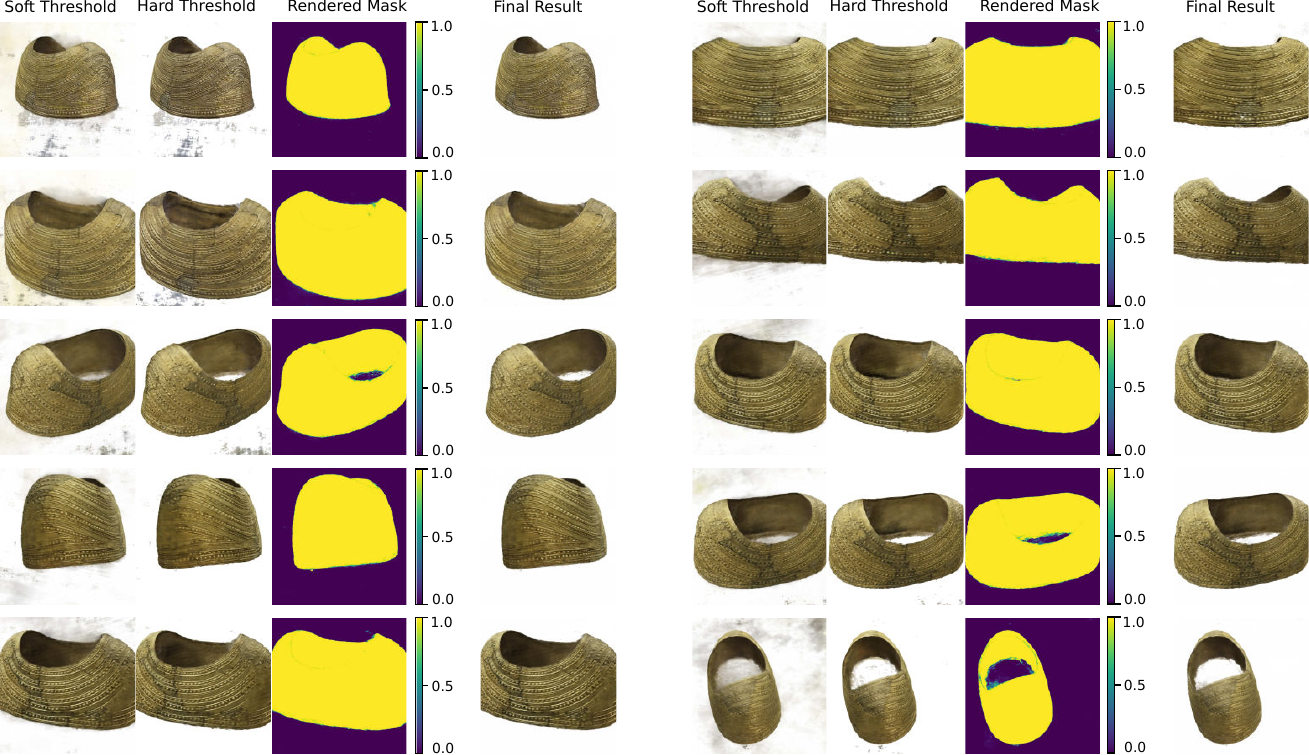}
   \caption{More examples of the effectiveness of our denoising approach for removing the background artifacts using rendered objectness probabilities which act as object masks on the goldcape scene~\cite{daniel_pett_gold_cape}.  }
   \label{fig:gold.cape.pipeline.long}
\end{figure}

As mentioned in section~\ref{sec:real.world.scenes}, we use a post-processing approach to denoise the rendered images of the objects. Noise is mostly caused by small particles that emerge in the training of the NeRF model and that blend in with the background in the training views, but become visible as the non-objects (including the background) are removed from the scene. In addition, for the points in the space that are inside of a dense object or behind the background, the objectness score is not trained well since the densities of the surface points block the training signals. All these reasons contribute to noisy renderings of the \ooi when using LaTeRF without additional denoising (see soft threshold results in Fig~\ref{fig:gold.cape.pipeline.long}). Our first step to mitigate this issue is to smooth the densities. Because the noisy particles are mostly steep `jumps' in density compared to the neighbouring points, substituting the value of every density with the average of its neighbours, including itself, will smooth these bumps. We repeat this averaging for $5$ steps. Afterward, we filter the points with densities lower than a small threshold (which was set to $0.2$ in the example in Fig~\ref{fig:gold.cape.pipeline.long}). We call this approach hard thresholding and it is evident in Fig~\ref{fig:gold.cape.pipeline.long} that it has helped to reduce the noise, but that there are still some unpleasant gray artifacts in the renderings. However, we do not directly use hard thresholding to render the RGB images of the object; we only use it to render the silhouette of the object to mask it out from the soft threshold results. After applying the hard threshold, we render the object mask by substituting the color with the objectness scores in Eq.~\ref{eq:discrete.object.rendering} and applying the sigmoid function:
\begin{equation}
    \label{eq:discrete.mask.rendering}
    \hat{P}_\text{obj}(r) = \text{Sigmoid}\Bigg(\sum_{i = 1}^{N} T^\text{obj}_i \big(1 - \exp (-\sigma_i p_i \delta_i)\big) s_i\Bigg).
\end{equation}
Note that we assume a background with $0$ objectness probability when rendering the object so that only object points with high densities will dominate this background and, as such, we obtain the object mask as an output. The rendered mask is then applied to the soft-threshold results to yield the rendered images of the object without background artifacts (final results are shown in Figure~\ref{fig:gold.cape.pipeline.long}). 

\section{Additional Real-world Results}

\begin{figure}[b]
  \centering
   \includegraphics[width=1\linewidth]{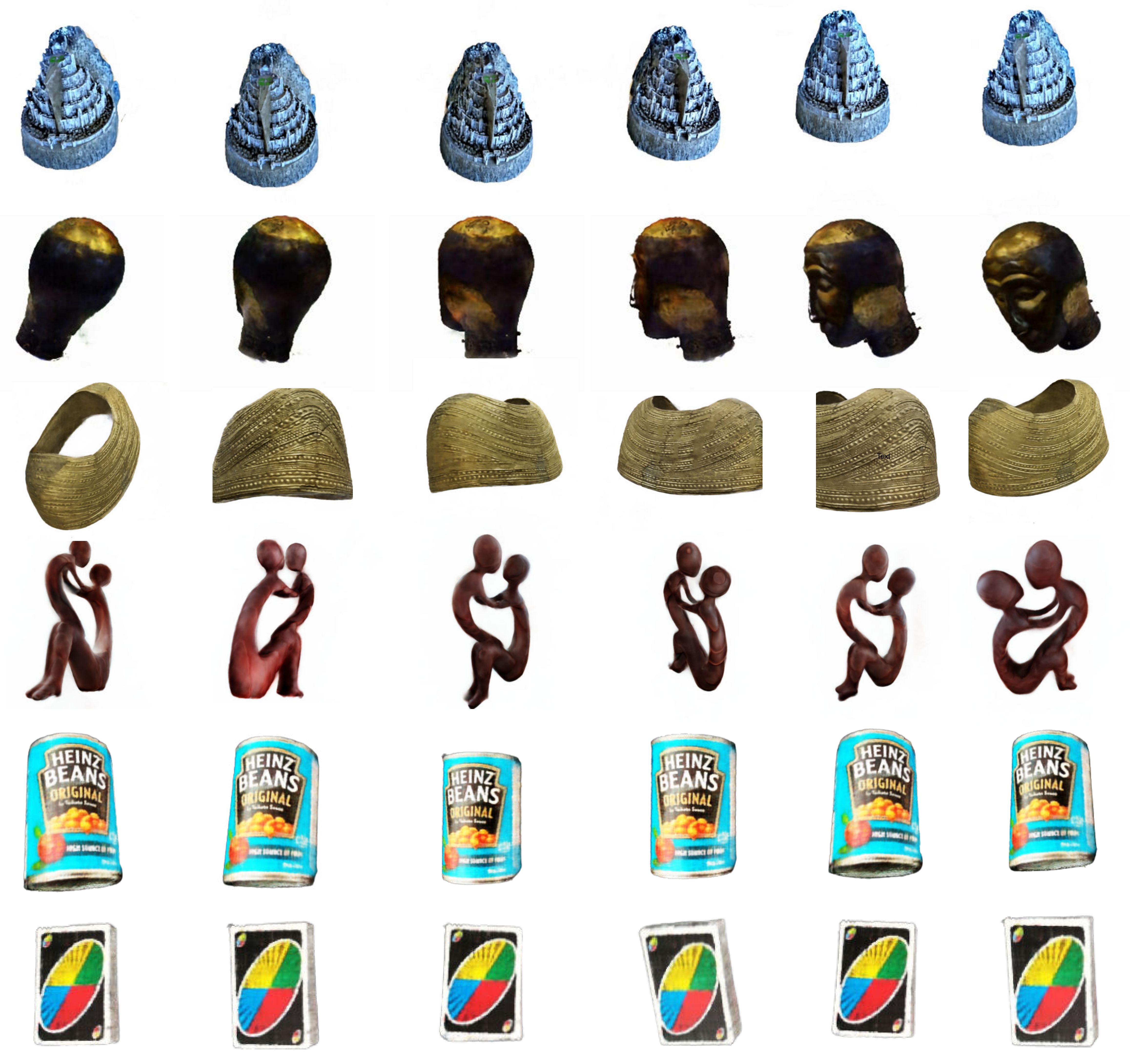}

   \caption{Additional real-world object extraction results. }
   \label{fig:additional.real}
\end{figure}

Novel-view renderings of additional objects (partly borrowed from \cite{mildenhall2020nerf,boss2021nerd}) are shown in Figure~\ref{fig:additional.real}. These examples include objects with detailed textures and geometries and objects with challenging shiny surfaces.

\section{Relighting the Object}

\begin{figure}[t]
\begin{floatrow}
\ffigbox{\caption{Relighting objects under three different illumination conditions.}}{\includegraphics[width=0.8\linewidth]{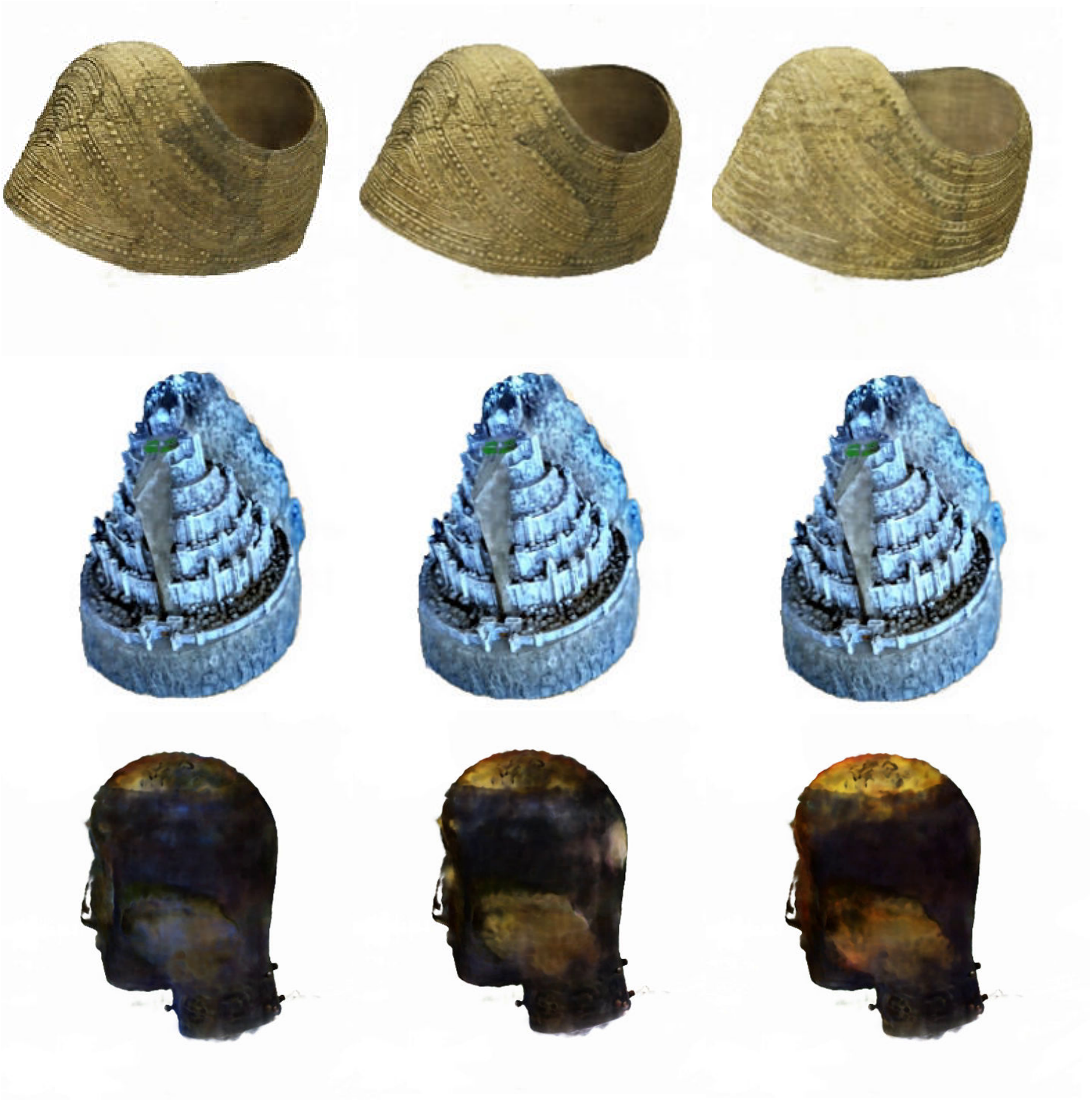}\label{fig:relighting}}
\ffigbox{\caption{An example of placing an extracted object in a scene with two different lighting conditions. }}{\includegraphics[width=\linewidth]{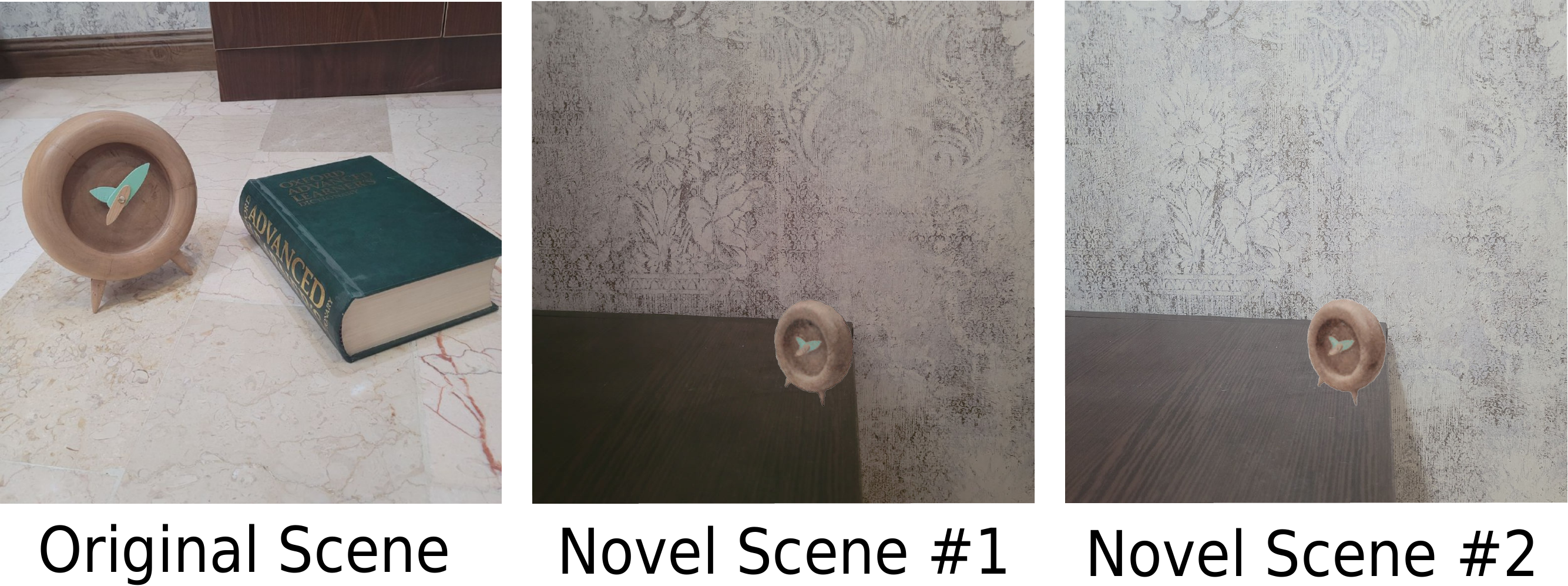}\label{fig:clock.in.novel.scenes}}
\end{floatrow}
\end{figure}

It is possible to leverage the dependence of color of a point on the view direction to `trick' the learned object radiance field to render the \ooi under novel lighting conditions~\cite{mildenhall2020nerf}. The view direction fed to the network to find the radiance (color) of points in the space can be manually rotated while keeping the camera fixed. The effect caused by this alteration is similar to changing the lighting of the scene, and it is possible to fit the novel lighting to be consistent with certain lighting conditions. Figure~\ref{fig:relighting} shows some of the real-world objects with three different illumination choices for a given, fixed view.



\subsection{Blending the Object in Novel Scenes}

The lighting setting can later be optimized with respect to the desired lighting in a novel scene, making an inserted object look consistent in a new scene. Figure~\ref{fig:clock.in.novel.scenes} shows an example of placing a 3D asset extracted by LaTeRF into a scene under two different illumination conditions.
\end{document}